\documentclass[conference]{IEEEtran}
\IEEEoverridecommandlockouts
\usepackage{cite,times}
\usepackage{amsmath,amssymb,amsfonts}
\usepackage{algorithmic}
\usepackage{graphicx}
\usepackage{textcomp}
\usepackage{xcolor}
\usepackage{amsmath,amssymb,amsfonts}
\usepackage{algorithm}
\usepackage{algorithmic}
\usepackage{graphicx}
\usepackage{textcomp,verbatim}
\usepackage{xcolor}
\usepackage{caption}
\usepackage{subcaption}
\usepackage{hyperref}
\usepackage[english]{babel}
\usepackage{amsthm}
\usepackage{amssymb}
\usepackage{parskip}

\begin{document}
	\title{Collisionless  Pattern Discovery in Robot Swarms Using 
		Deep Reinforcement Learning
	}
	
	\author{\IEEEauthorblockN{Nelson Sharma\IEEEauthorrefmark{1},
			Aswini Ghosh\IEEEauthorrefmark{1}, Rajiv Misra\IEEEauthorrefmark{1}, Supratik Mukhopadhyay\IEEEauthorrefmark{2}, and Gokarna Sharma\IEEEauthorrefmark{3}}\\
		\IEEEauthorblockA{\IEEEauthorrefmark{1}Department of Computer Science and Engineering, Indian Institute of Technology, Patna, India
			\IEEEauthorblockA{\IEEEauthorrefmark{2}Department of Environmental Sciences, Louisiana State University, Baton Rouge, Louisiana, USA 
				\IEEEauthorblockA{\IEEEauthorrefmark{3}Department of Computer Science, Kent State University, Kent, Ohio, USA\\
					E-mail: \{nelson\_2121cs07, aswini\_2121cs02, rajivm\}@iitp.ac.in; mmukho1@lsu.edu; gsharma2@kent.edu} }
		}
	}
	
	\maketitle
	
	\begin{abstract}
	    We present a deep reinforcement learning-based framework for automatically discovering patterns available in any given initial configuration of fat robot swarms. In particular, 
		we model the problem of collision-less gathering and mutual visibility in fat robot swarms  and discover patterns for solving them using our framework. 
		We show that by shaping reward signals based on certain constraints like mutual visibility and safe proximity, the robots can discover collision-less trajectories leading to ``well-formed" gathering and visibility patterns. 
	\end{abstract} 
	


\section{Introduction}
	In recent years, swarm robotics \cite{dorigo2014swarm} has gained a lot of attention from the research community. In a swarm, a  number of   robots work together collectively for performing a  certain task, such as disaster recovery, foraging, mapping, etc \cite{blum2008swarm}. In many cases, it is assumed that there is no explicit means of communication between the robots (e.g., in a bandwidth-limited disaster zone); a robot can only see others through a camera that provides a $360^{\circ}$ view. The robots are {\em anonymous} (no unique identifiers), {\em disoriented} (no agreement on coordinate systems or units of distance), {\em autonomous} (no external control), and {\em indistinguishable} (no external identifiers). In the {\em classical oblivious swarm} model \cite{b2,b50},  the position of every robot is considered as a point on 2D plane and has a local co-ordinate system. Each robot senses other robots using  sensory equipment installed on it,  such as a camera. 
	However, the robots are considered {\em silent} since it is assumed that there is no direct communication among them \cite{b2,b50}. Each robot executes the same algorithm.
	
	Additionally, each robot executes {\em Look-Compute-Move} (LCM) cycle -- when a robot is activated, it takes a snapshot of its surroundings from its local perspective ({\em Look}), then it uses the snapshot and the algorithm it is executing to determine an action to take ({\em Compute}), and finally it performs the action ({\em Move}). A swarm of robots can operate   asynchronously, semi-synchronously or fully synchronously. In the {\em fully synchronous} operating model, every  robot in the swarm becomes active in every LCM cycle and cycles of every robot are synchronized (i.e., there is a global clock).
	In the {\em semi-synchronous} operating model, not all swarm robots  necessarily become active in every LCM cycle, but LCM cycles are synchronized (i.e., there is still a global clock). In the {\em asynchronous} operating model, not every robot is necessarily active all the time, the duration of each LCM cycle is arbitrary, and the LCM cycles of different robots are not synchronized (i.e., no global clock).
	In the classical model, robots are oblivious because they do not keep memory of what they did in  any past LCM cycle. 
	
	
	
	There has been a lot of research recently in forming patterns  with maximum visibility and  gathering without collisions for swarm robots in the classical model \cite{b5,b6,b7,b8,b9,b10,b11,b12}. {\em Pattern formation} concerns arranging $n$  robots in a given pattern  on the plane. A configuration of robots matches the pattern if rotation, translation, and scaling of the configuration  can produce the pattern explicitly \cite{b5,b6,b7,b8,b9,b10,b11,b12}. An algorithm solves a pattern formation problem, if given any  configuration of $n$ robots located initially at distinct points on a plane, it enables them to reach a configuration that matches the given pattern and remain stationary thereafter \cite{b37}. 
	Specific patterns like {\em circle formation} have been considered in \cite{b21,b22,b23,b24,b25}. {\em Convex hull} and {\em cubic spline}  formations as  solutions to the {\em mutual visibility} problem -- i.e., each robot sees all others in the swarm,  have also been studied in the classical model \cite{b15,b39}. 
	Pattern formation has been studied in \cite{b40} for  UAVs. Pattern Formation  via Deep Reinforcement Learning (DRL) has been studied in \cite{b41}.
	The objective in the  {\em gathering} problem is to arrange the robots as close together as possible around a  gathering point (that may or may not be known a priori) \cite{b36}.

	Existing pattern formation algorithms belong to two broad classes: {\em combinatorial} algorithms and {\em machine learning-based} approaches. In the combinatorial paradigm \cite{b1,b2,b3,b4,b5,b6,b7,b8,b9,b10,b11,b12,b13,b14,b15,b16,b17,b18,b19,b20,b21,b22,b23,b24,b25,b26,b27,b28,b29,b30,b31,b32,b33,b34,b35}, it is assumed that the target pattern to be formed is given as input to the robots and one designs a sophisticated algorithm to arrange robots to be positioned on the target pattern positions, starting from a given initial configuration. Designing such algorithms is labor-intensive, error-prone, and time consuming. Additionally, the designed combinatorial algorithm may not work for any initial configuration, for example, many combinatorial algorithms on the literature assume that the  initial configurations are {\em asymmetric} \cite{b1,b2,b3,b4}.  The same kind of assumption of asymmetric initial configuration extends also to the combinatorial algorithms for gathering in the literature.  Existing machine learning based approaches \cite{b36,b37,b38,b39,b40,b41,b42,b43,b44,b45,b46,b47,b48,b49} attempt to solve pattern formation for a given target pattern, i.e., similarly as in the combinatorial algorithms, the target pattern needs to be given to the machine learning algorithm.  
	
	It would be ideal, given any initial configuration (symmetric/asymmetric), if the valid target patterns could be identified and discovered, even when the target pattern is not given as input a priori. In this paper, we precisely aim to do that for the first time in the context of swarm robotics.     
	Particularly, we seek a \emph{single algorithm} that can {\em identify} and \emph{discover} patterns that provide solutions to problems such as mutual visibility or gathering, rather than giving as input the solution pattern beforehand and the goal of the algorithm is to reach to the given target from the given initial configuration. 

	
	We present a DRL-based framework for automatically discovering patterns available in any given initial configuration of robot swarms. We depart from the literature and consider robot swarms which are not points but {\em fat} -- and opaque robots with an extent and hence they occupy certain area. In particular,  we model the problem of collision-less gathering and mutual visibility in fat opaque robot swarms  and discover patterns for solving them using our framework. Since our proposed approach is DRL-based, it is based on designing good reward functions. We show that designing  reward signals based on  constraints like mutual visibility and safe proximity, the robots can discover collision-less trajectories leading to ``well-formed" gathering and visibility patterns. 
	
	In summary, 
	we make the  following two major  contributions. 
	\begin{itemize}
		\item We model the problem of automatic pattern discovery for collision-less gathering in robot swarms
             and formulate it in terms of DRL framework.

		\item We propose DRL based approach for the collision-less gathering and pattern discovery using multiple policies to discover ``well-formed'' gathering patterns.
		
	\end{itemize}

	\section{RELATED WORK}

 Specific target pattern formation for swarm robots have been extensively studied in \cite{b38,b39,b25,b26,b27,b28,b29}. Optimal convex hull pattern formation assuming obstructed visibility has been studied in \cite{b38}. In \cite{b39}, the authors studied proposed self organised cubic spline based pattern formation method. Several researchers have  studied circular pattern formation \cite{b25} problem. In \cite{b27}, authors have investigated the plane formation problem that requires a swarm of robots moving in the three-dimensional Euclidean space to land on a common plane and considered  the fully synchronous setting  and where robots are endowed with only visual perception.
 Most works that studied pattern formation problem considered the classic oblivious point swarm  model \cite{b28,b37}. 
 In \cite{b37}, the  authors presented a combinatorial framework for the pattern formation problem and used a randomized procedure for leader election in their proposed algorithm. They measured the time complexity of their algorithms in “epochs” (i.e., time interval in which every robot performs at least one look-compute-move step).	

For the fundamental problem of gathering, the authors in \cite{b14} considered gathering of $N$ autonomous robots on a plane, which requires all robots to meet at a single point  under {\em limited visibility}. Here the point of gathering is not known beforehand. The authors in \cite{b36} significantly improved the time complexity of gathering under limited visibility model with the assumption that the   point of gathering is known beforehand.  

The authors in \cite{b35} considered the state machine representation to formulate the gathering problem and developed a distributed algorithm that solves the problem of gathering for any number of fat robots. 
%
  In \cite{b41},  the authors  proposed a Deep Reinforcement Learning (DRL)-based method that generates general-purpose pattern formation strategies for any target pattern.
	%
 A bio-inspired algorithm was proposed in \cite{b45} to control robot swarms for pattern formation. The authors in \cite{b47} have  formulated reward functions to ensure  drones arrive at their destinations following predefined routes avoiding collisions. The authors in \cite{b48} presented the idea of guided policy learning and investigated how to learn to control a group of cooperative agents with limited sensing capabilities such as robot swarms. The authors in \cite{b49} presented a framework  for collision-less flying of drones over a directed circulant digraph.
	
However, none of the above discussed approaches  have attempted discovering collision-less patterns where mutual visibility is maximized with or without predefined gathering points. 
	%
	%
In this paper, we consider a DRL-based framework  for discovering ``well-formed'' patterns for mutual visibility and gathering where gathering point may not be known beforehand.

		\section{System Model and Preliminaries}

\begin{figure}[!t]
		\centering
		\includegraphics[width=4.2cm]{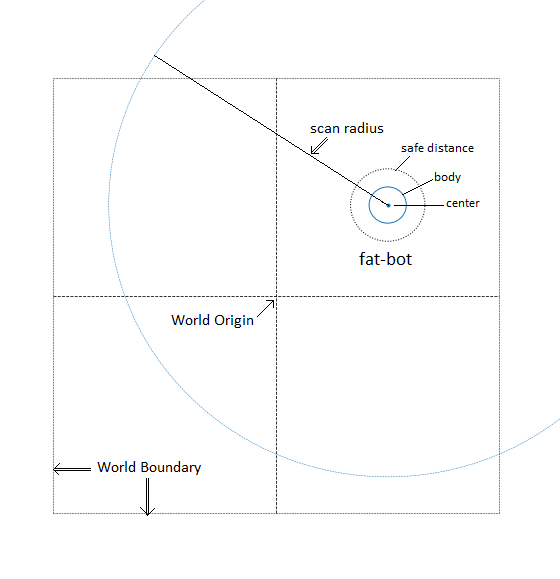}
		\includegraphics[width=4.5cm]{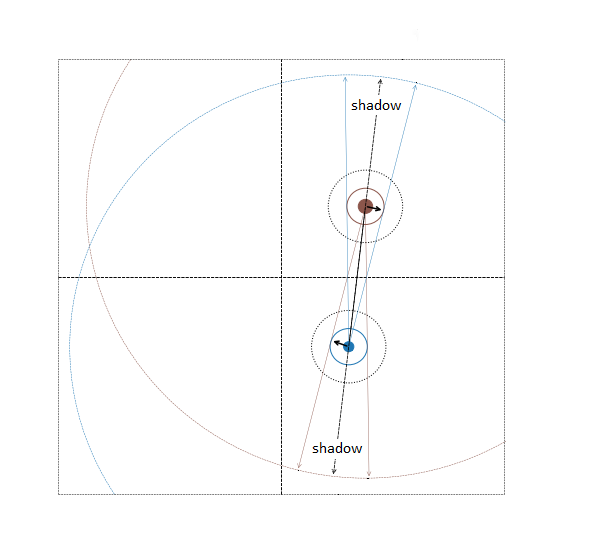}
	
		\vspace{-6mm}
		\caption{(left) Fat robot model (right) Visibility model}
		\vspace{-6mm}
		\label{fig:world}
	\end{figure}

	\noindent{\bf Robot Swarm.}
	A robot swarm is a collection on $n$ robots moving autonomously inside a rectangular region of size $(2X_{w}, 2Y_{w})$ on a 2D Cartesian plane with origin being at the center of the rectangle. The robots are \emph{fat} and \emph{opaque}, meaning that a robot $(n \in N)$ is represented as a circular opaque disk of radius $R_{bot}$. As shown in Fig.~ \ref{fig:world} (left), a robot has a center and a circular body of fixed size. A robot's position vector $(P_{n})$ and velocity vector $(V_{n})$ are assumed to be known at all times, recorded using sensors internal to robots. It is assumed that all robots are fitted with sensors that can detect the positions of neighboring robots within a fixed distance from their centers, called the \emph{scan-radius} ($R_{scan}$). The sensors' readings for a robot provide: (i)  the number of  neighbors ($G^{all}_{n}$), (ii) the number of fully visible neighbors ($G^{vis}_{n}$), and (iii) the number of occluded neighbors ($G^{occ}_{n}$). Additionally, a \emph{safe distance} ($\delta_{s}$) is defined, which can be used to calculate a measure of how unsafe is the current positioning of the robots.
	
	Since robots are fat and opaque, a robot $B_{i}$ is said to be {\em occluded} by robot $B_{j}$ from the view of robot $B_{k}$ if any part of $B_{i}$ lies in the shadow formed by $B_{j}$ assuming that a point source of light placed at the center of $B_{k}$, as shown in Fig.~\ref{fig:world} (right). Two robots are \emph{mutually visible} to each other if and only if none of them are occluded by any robot from the other's view. The robots are allowed to move in any direction by changing their velocity vector $(V_{n} = [V_{n}^{x}, V_{n}^{y}] )$ within a fixed range ($V^{min} \le V_{n}^{x}, V_{n}^{y} \le V^{max}  $ ) at each time cycle.

	\noindent{\bf Problem Formulation.}
	The goal of robot swarm is to discover appropriate patterns for gathering such that the robots do not collide with each other and maintain maximum visibility among one another during pattern formation. We consider two gathering problem variations as follows:
	
	\noindent{\bf \em Gathering at a Predefined Point.}\label{target}
	In this case, the robots are initially scattered far apart, i.e., mostly outside the scan-radius of each other. It is assumed that the gathering point is at the origin of the environment.
	
	\noindent{\bf \em Gathering without a Predefined Point.}\label{notarget}
	We assume a very large scan-radius and all robots are within the scan-radius of each other. Gathering point is not known a priori and the robot swarm has to decide the appropriate gathering place.
	
	\noindent{\bf Markov Decision Process.}
	We formulate the above variations of the gathering problem  as an appropriate \emph{Markov Decision Process (MDP)} and solve it using a DRL-based approach that tries to maximize its underlying reward function. The MDP is formulated as follows:

	\begin{itemize}
		\item {\em State Space} ($\mathcal{S}$): The State Space contains the position coordinates of all the robots.
		\begin{equation}
			\mathcal{S} =  \{P_{n} \} : \forall n \in \{1...N \}
		\end{equation}
		
		\item {\em Action Space} ($\mathcal{A}$): The Action Space contains the velocity components that can be set at a given time-step, for all the robots.
		\begin{equation}
			\mathcal{A} =  \{[V_{n}^{x}, V_{n}^{y}] \} : \forall n \in \{1...N \}
		\end{equation}

		\item {\em Reward Function} ($\mathcal{R}$): The reward for each time-step is based on the \emph{Reward Signal} ($\mathcal{C}$) from the current state ($S_{t}$) and the next state ($S_{t+1}$).
		\begin{equation}
			\mathcal{R}(S_{t+1}|S_{t}, A_{t}) =  \mathcal{C}(S_{t+1}) - \mathcal{C}(S_{t})
		\end{equation}
		where, the Reward Signal for a given state is a weighted sum of multiple signals. We define two reward signals, $\mathcal{C}_{1}$ and $\mathcal{C}_{2}$ as follows:
		\begin{equation}
			\mathcal{C}_{1}(S_{t}) =  \sum_{C \in C_{1}} C(S_{t})
		\end{equation}
		\begin{equation}
			\mathcal{C}_{2}(S_{t}) =  \sum_{C \in C_{2}} C(S_{t}) 
		\end{equation}
		
		where, $C_{1}$ and $C_{2}$ are sets of signals defined as:
		
		\begin{equation}
			C_{1} = \{ C_{close}, C_{safety}, C_{neighbors}, C_{visible} \}
		\end{equation}
		\begin{equation}
			C_{2} = \{ C_{nclose}, C_{safety}, C_{neighbors}, C_{visible} \}
		\end{equation}
		
		where, the members are defined as follows:
		\begin{itemize}
			\item $C_{close}$ : measure of closeness of robots from the origin. This signal causes robots to move closer to the gathering point, i.e., the origin.
			
			\begin{equation}
				C_{close}(S_{t}) = 1 - W_{close} \sum_{n \in N}|| P_{n}(t) ||
			\end{equation}
			\item $C_{safety}$ : is the sum of safety measure ($F_{safe}$) for all robots, which helps robots to avoid collisions  maintaining appropriate distance between each other.
			\begin{equation}
				C_{safety}(S_{t}) = W_{safety} \sum_{i,j \in N, i\ne j} F_{safe}(i,j, t) 
			\end{equation}
			where, $F_{safe}$ is defined between robots $i$ and $j$ at time-step $t$, based on a known constant \emph{safe-distance} ($\delta_{s}$), as follows:
			\[ 
			F_{d} = ||P_{i}(t)-P_{j}(t)|| - \delta_{s}
			\]
			\[   
			F_{safe}(i,j, t)  = 
			\begin{cases}
				\sqrt{ \delta_{s} + R_{bot} - F_{d}}, &F_{d} \ge 0, \\
				\text{0}, &F_{d} < 0, \\
			\end{cases}
			\]
			
			\item $C_{neighbors}$ : denotes the number of neighbors and encourages the robots to have more robots in their neighborhood.
			\begin{equation}
				C_{neighbors}(S_{t}) = W_{neighbors} \sum_{n \in N} G_{n}^{all}(t) 
			\end{equation}
			
			\item $C_{visible}$ : denotes total number of neighbors that are currently fully visible for each robot and encourages maximum visibility within the robot swarm.
			
			\begin{equation}
				C_{visible}(S_{t}) = W_{visible} \sum_{n \in N} G_{n}^{vis}(t) 
			\end{equation}
			
			\item $C_{nclose}$ : measures the average distances from all neighbors for each robot. In the absence of a priori gathering point, this signal encourages robots to move closer to each other for  gathering.
			\begin{equation}
				C_{nclose}(S_{t}) = 1 - W_{nclose} \sum_{i,j \in N, i \ne j}  ||P_{i}(t)-P_{j}(t)|| 
			\end{equation}
			
			The weights assigned to each signal ($W_{i}$) are chosen so as to normalize the values to the range $[0,1]$.
		\end{itemize}
		
		\item {\em Initial States} ($\mathcal{S}_{0}$): We consider three types of initial state configurations:
		\begin{itemize}
			\item \emph{Scattered}: robots are uniformly scattered along a roughly circular shape with radius $\min(X_{w}, Y_{w})$.
			
			\item \emph{Distributed}: robots are equally scattered near any two of the corners of the environment.
			
			\item \emph{Packed}: robots are closely placed in a line (or multiple lines).
			
		\end{itemize}
	An initial state configuration $(\mathcal{S}_{0}, \epsilon)$ is chosen at the start of each episode where $\mathcal{S}_{0}$ is a set of $n$ coordinates, one for each robot, and $\epsilon$ is the noise radius. A robot $(B_{n})$ is placed at a location ($P_{n}(0)$) given by:
    \begin{equation}
        P_{n}(0) = P_{n}^{0} + \mathcal{U}(-\epsilon, \epsilon)
    \end{equation}
    where $P_{n}^{0}$ is an predefined point generated based on the initial state $(\mathcal{S}_{0})$. Initial position of a robot is determined by adding uniformly sampled noise ($\epsilon$) to each dimension of $P_{n}^{0}$.
		
	\end{itemize}	
	
	For gathering at a predefined point, we use the reward signal $\mathcal{C}_{1}$, which tries to maximize the $C_{close}$ signal causing robots to move closer to a pre-defined gathering point. In contrast,  for gathering without a predefined point, we shall use the reward signal $\mathcal{C}_{2}$ in our MDP which tries to maximize the $C_{nclose}$ signal causing robots to move closer to each other. For both  the problems, the common reward signal is $\mathcal{C}_{safe}$ which encourages collision-less movement at a safe distance; and $\mathcal{C}_{visible}$ which encourages maximum mutual visibility.

	\noindent{\bf Proximal Policy Optimization (PPO).}
	PPO \cite{b51} is a policy optimization algorithm where the parameterized policy is updated using one of two objective functions. The first is \emph{clipped objective} function ($L_{C}$) defined as follows:
	\begin{equation}
		L_{C} =  \mathbb{E}_{t} \left[ \min(r_{t}\hat{A}_{t}, clip(r_{t}, 1-\epsilon_{c}, 1+\epsilon_{c})\hat{A}_{t}) \right]
	\end{equation}
	where, $\hat{A}_{t}$ represents the advantage estimate, $\epsilon_{c}$ represents the \emph{clipping hyper-parameter} and $r_{t}$ is the ratio of probabilities given by the current policy ($\pi_{i}$) and an older version of it ($\pi_{i-1}$), defined as:
	\begin{equation}
		r_{t} = \frac{\pi_{i}(A_{t}| S_{t})}{\pi_{i-1}(A_{t}| S_{t})} 
	\end{equation}
	The clipped objective can be maximized using a mini-batch ($\mathcal{D}$) of trajectories, in which case the objective becomes:
	\begin{equation}
	L_{C} = \frac{1}{|\mathcal{D}|T}\sum_{\mathcal{D}} \sum_{t=0}^{T} \min(r_{t}\hat{A}_{t}, clip(r_{t}, 1-\epsilon_{c}, 1+\epsilon_{c})\hat{A}_{t})
\end{equation}
where, $T$ represents the trajectory length. The advantage is estimated using a truncated version of {\em Generalized Advantage Estimation} (GAE)\cite{b52} defined as:
\begin{equation}
	\delta_{t} = \mathcal{R}_{t} + \gamma V(S_{t+1}) - V(S_{t}) 
\end{equation}
\begin{equation}
	\hat{A}_{i}(\gamma, \lambda) = \sum_{l=0}^{T-t-1} (\gamma\lambda)^{l}\delta_{l+t}
\end{equation}
where, $\lambda$ is a hyper-parameter, $\gamma$ is the discount factor, and $V$ represents the value function.

Notice that PPO is an on-policy algorithm, i.e., it uses the latest policy to sample actions during the learning process. The exploration is dependent on the randomness in the stochastic policy. As the policy is tuned over time, the randomness in sampled action decreases which often causes the agent to explore less, leading to the policy settling in a local optima. We shall take advantage of this fact and use multiple policies to move around local optima until we find a satisfactory one. This can be achieved, in one way, by altering the value of discount factor during multiple runs of the learning algorithm.
	
	\section{Algorithm}
	To discover a gathering pattern, we use Algorithm \ref{ALGORITH_2} to generate two policies sequentially using an augmented version of the Clipped Proximal Policy Optimization (Clipped-PPO) algorithm developed in \cite{b51}. The first policy ($\Pi_{0}$) is called the \emph{Base} policy and the second one ($\Pi_{1}$)  is called the \emph{Auxiliary} policy. The base policy finds near-optimal patterns in most cases but may sometimes get struck on sub-optimal patterns (i.e.,  {\em local optima}). The Auxiliary policy is used to break out of local optima and improve the reward further. The base policy uses a larger value of the discount factor ($\gamma$) and trains on longer horizons. On the other hand, the auxiliary policy uses a smaller value of discount factor and trains on short horizons using only the sub-optimal patterns formed by the base policy as the initial states. The pattern can be sub-optimal, in the sense that  the  robots do satisfy the gathering criteria only to a certain degree. In other words, it might be possible that the reward can be further maximized to obtain a more optimal pattern. In such a case, the auxiliary policy is trained starting at various sub-optimal patterns formed by the base policy and is expected to improve the pattern further using a few more steps. The algorithm additionally uses hyper-parameters which are described in Table \ref{tab:hyper}.

\begin{table}[ht]
\begin{center}
\caption{Hyper-parameters used in proposed algorithms}
\begin{tabular}{ |c|c| }
\hline
Hyper-Parameter & Description \\
\hline
$\gamma$ & Discount Factor  \\
\hline
$\lambda$ & GAE Lambda \\
\hline
$\epsilon_{c}$ & Clipping Range for Objective  \\
\hline
$\alpha$ & Learning Rate for policy updates  \\
\hline

$\beta$ & Learning Rate for value updates   \\
\hline
$Z$ & No. of Learning Epochs  \\
\hline
\end{tabular}
\label{tab:hyper}
\end{center}
\end{table}

	\begin{algorithm}
	{\footnotesize
		\caption{ Gathering using Clipped PPO }
		\label{ALGORITH_1}
		\begin{algorithmic}[1]
			\REQUIRE Set of Initial States $({\sigma})$
			\REQUIRE Hyper-Parameters $( \gamma, \lambda, \epsilon_{c}, \alpha, \beta, Z)$
			
			\STATE Initialize Policy-Network $(\pi_{\theta})$ and Value-Network $(V_{\phi})$ with parameters $\theta_{0}$ and $\phi_{0}$ respectively
			
			\STATE Initialize Learning Environment with a uniform  Initial State Distribution on set $\sigma$.
			
			\STATE Initialize an empty set $\sigma^{*}$
			\FOR{$i = 1$ \TO $Z$ }
			\STATE Collect batch of trajectories $\mathcal{D}_{i} = \{S_{t}, A_{t}, R_{t}, S_{t+1}\}$ from the environment using current policy $\pi_{\theta_{i}}$ and compute \emph{Rewards-to-go} ($\hat{R}_{t}$) at each time-step $t$
			\[
			\hat{R}_{t} = \sum_{l=t}^{T} \mathcal{R}(S_{l+1}|(S_{l}, A_{l}))
			\]			
			\STATE Compute Clipped Objective ($L_{C}$) for batch $\mathcal{D}_{i}$ using equation (16)
			
			\STATE Update Policy Network using:
			\[
			\theta_{i+1} = \theta_{i} + \alpha \cdot \nabla_{\theta} L_{C}
			\]
			
			\STATE Update Value Network using:
			\[
			E(\phi_{i}) = \frac{1}{|\mathcal{D}_{i}|T}\sum_{\mathcal{D}_{i}} \sum_{t=0}^{T} (V_{\phi_{i}}(S_{t}) - \hat{R}_{t})^{2}
			\]
			\[
			\phi_{i+1} = \phi_{i} - \beta \cdot \nabla_{\phi} E(\phi_{i})
			\]
			\ENDFOR
			\STATE Evaluate Policy $\pi_{\theta}$ in the Environment using initial states from $\sigma$. For each $S_{i} \in \sigma $, add the final state $S^{*}_{i}$ to set $\sigma^{*}$ only if $\mathcal{S}^{*}_{i}$ is not a terminal state.
			
			\STATE Output policy $\pi_{\theta}$ and set of states $\sigma^{*}$
			
		\end{algorithmic}
		}
	\end{algorithm}

	The clipping parameter ($\epsilon_{c}$) is gradually increased with each epoch of Algorithm \ref{ALGORITH_1}. The Discount Factor Schedule ($F_{\gamma}$) is a function which provides appropriate value of discount factor on each successive run of Algorithm \ref{ALGORITH_2}. We train only two successive policies where the value of Discount Factor is high for the first run and reduced during the second run. The auxiliary policy is responsible for improving patterns discovered by the base policy and can be trained on customized reward signals that give rise to various patterns while satisfying given gathering criteria.
	
	\begin{algorithm}
	{\footnotesize
		\caption{ Pattern Discovery for robot swarm }
		\label{ALGORITH_2}
		\begin{algorithmic}[1]
			\REQUIRE Initial State Configuration $(\mathcal{S}_{0}, \epsilon)$
			\REQUIRE Hyper-Parameters $( \lambda, \epsilon_{c}, \alpha, \beta, Z)$
			\REQUIRE Discount Factor Schedule $( F_{\gamma} ) $
			
			\STATE Initialize set of initial states $\sigma \gets (\mathcal{S}_{0}, \epsilon)$
			
			\STATE Initialize $t \gets 0$
			
			\STATE Initialize empty list of policies ($\Pi$)
			
			\WHILE{$|\sigma|>0$ }
			
			\STATE Run Algorithm[1] with Set of Initial States ($\sigma$) and hyper-parameters $( F_{\gamma}(t), \lambda, \epsilon_{c}, \alpha, \beta, Z)$ to obtain $\pi^{t}$ and $\sigma^{t}$ 
			
			\STATE Append $\pi^{t}$ to list of policies $\Pi$
			
			\STATE Update set of initial states $\sigma \gets \sigma^{t}$
			
			\STATE $t \gets t+1$

			\STATE Evaluate policy list ($\Pi$) with given Initial State Configuration $(\mathcal{S}_{0}, \epsilon)$ to obtain desired trajectory set $\mathcal{T}$
			\STATE If $\mathcal{T}$ forms valid collision-less patterns then Break
			\ENDWHILE
			\IF {$|\sigma|=0$}
			\STATE No valid pattern formation trajectory was discovered
			\ELSE
			\STATE Output Policy List $\Pi$
			\ENDIF 
		\end{algorithmic}
		}
	\end{algorithm}
	
	\section{Experiments and Results}
	
	To test the pattern discovery algorithm (Algorithm \ref{ALGORITH_2}), we conducted a total of six experiments under three different swarm compositions with two different gathering criteria (reward signals) starting in one of the three initial state configurations (Scattered, Distributed, Packed). The experimental settings are described in Table \ref{tab:T_exp_settings}.
	
	\begin{table}[ht]
		\begin{center}
			\caption{Experiment Settings}
			\begin{tabular}{ |c|c|c|c| } 
				\hline
				Experiment & Swarm Size &  Gathering Target & Initial States\\ 
				\hline
				A & $6$ & Origin & Packed \\ 
				\hline
				B & $8$ & Origin & Scattered \\ 
				\hline
				C & $10$ & Origin & Distributed \\ 
				\hline
				
				D & 6 & Undefined & Packed \\ 
				\hline
				E & 8 & Undefined & Scattered \\ 
				\hline
				F & 10 & Undefined & Distributed \\ 
				\hline
			\end{tabular}
			\label{tab:T_exp_settings}
		\end{center}
	\end{table}
	
\begin{figure}[!t]
	\vspace{-4mm}
		\centering
		\includegraphics[width=6cm]{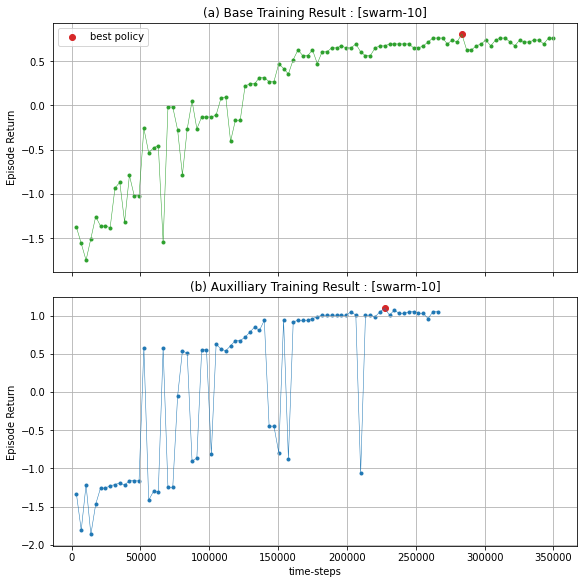}
		\caption{[Exp C] Training result for Base Policy and Auxiliary Policy for swarm of size 10 with pre-defined gathering point  }
		\label{fig:train_c_1}
		\vspace{-4mm}
	\end{figure}

\begin{figure}[!t]
		\centering
		\includegraphics[width=6cm]{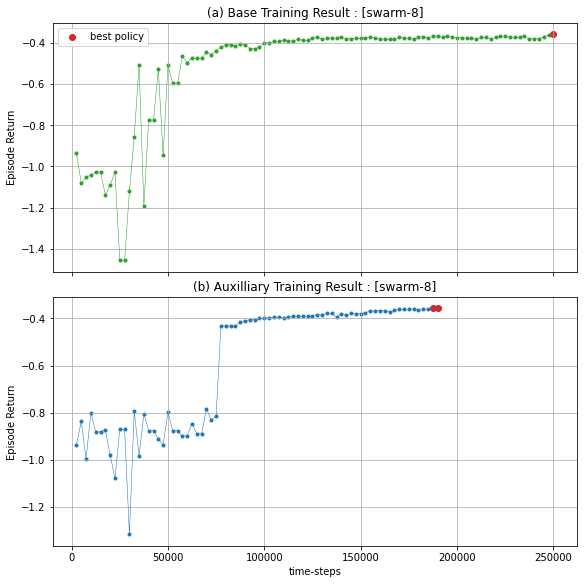}
		\caption{[Exp E] Training result for Base Policy and Auxiliary Policy for swarm of size 8 with undefined gathering point  }
		\label{fig:train_e_1}
		\vspace{-4mm}
	\end{figure}

	The training results for swarm size of 10 and 8 (in experiments C and E) are shown in Figs. \ref{fig:train_c_1} and  \ref{fig:train_e_1}, respectively. Note that, in both cases, the auxiliary policy improves the patterns formed by the base policy which is evident from the improvement seen in the reward signal during auxiliary training.

\begin{figure}[htp]
	\vspace{-3mm}
		\centering
		\includegraphics[width=8cm]{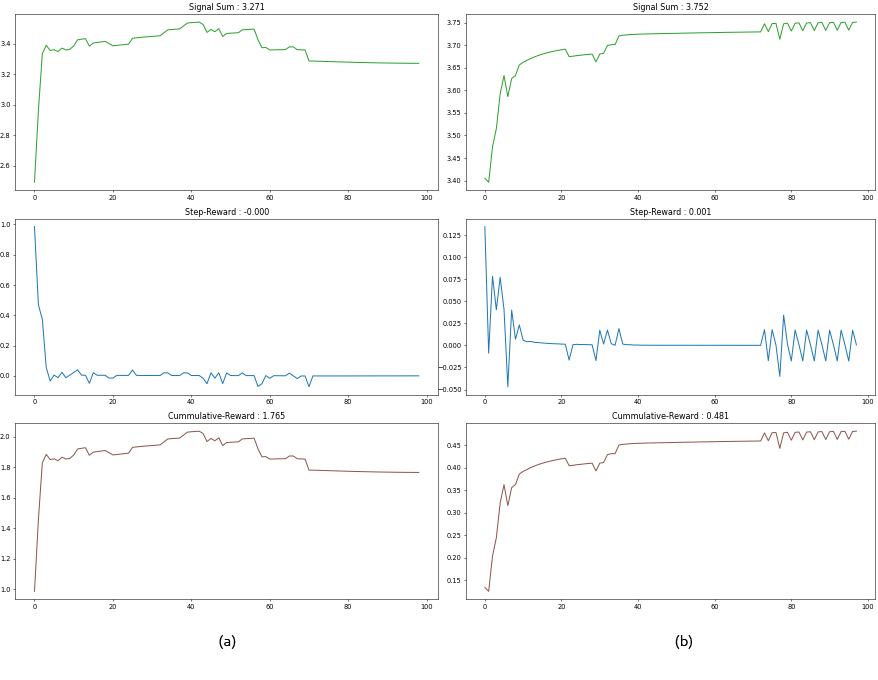}
		\vspace{-3mm}
		\caption{[Exp C] Cumulative Reward obtained during single episode: (a) Reward Obtained by Base Policy, (b) Reward Obtained by Auxiliary Policy.  }
		\label{fig:testrun}
		\vspace{-6mm}
	\end{figure}

	Fig.~ \ref{fig:testrun} shows the reward obtained by robot during a single episode of pattern formation. When the \emph{step-reward} received by the agent becomes zero, it indicates that the base policy cannot improve the pattern any further, and hence, has converged to a stable pattern. It can be seen that the base policy leads to a sub-optimal pattern which is then improved by the auxiliary policy.
	
\begin{figure}[!t]
		\centering
		\includegraphics[width=8cm]{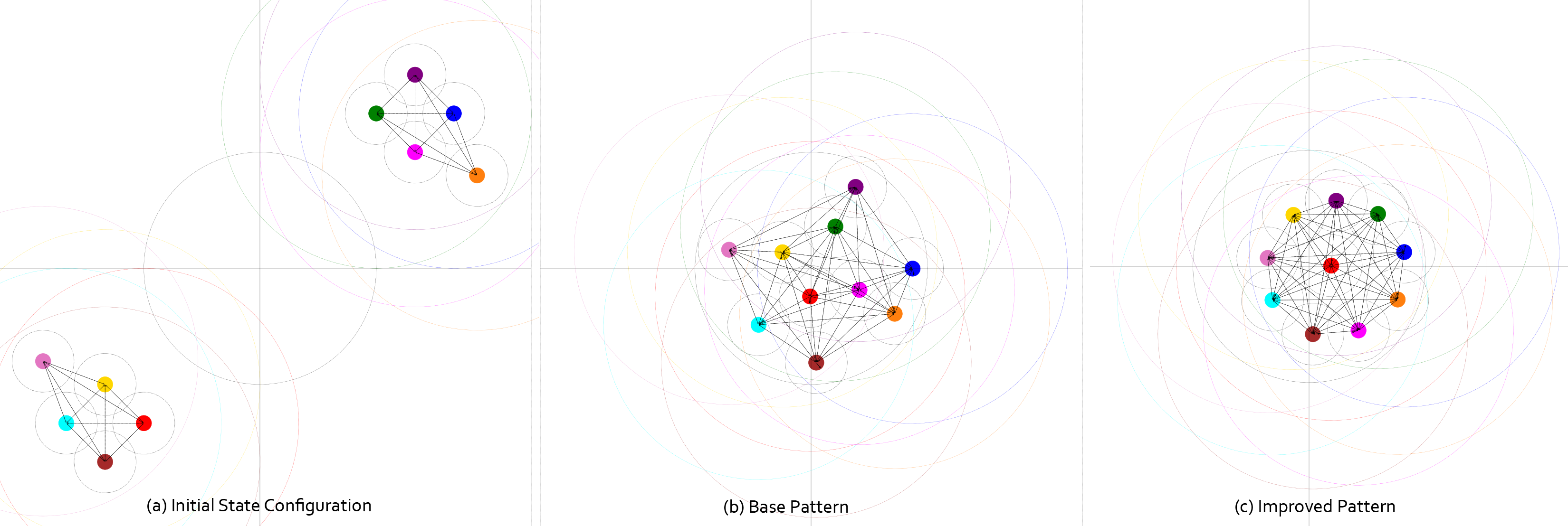}
		\caption{[Exp C] Pattern formation for swarm of size 10 in distributed initial state with pre-defined gathering point  }
		\label{fig:res10}
		\vspace{-4mm}
	\end{figure}

\begin{figure}[!t]
		\centering
		\includegraphics[width=8cm]{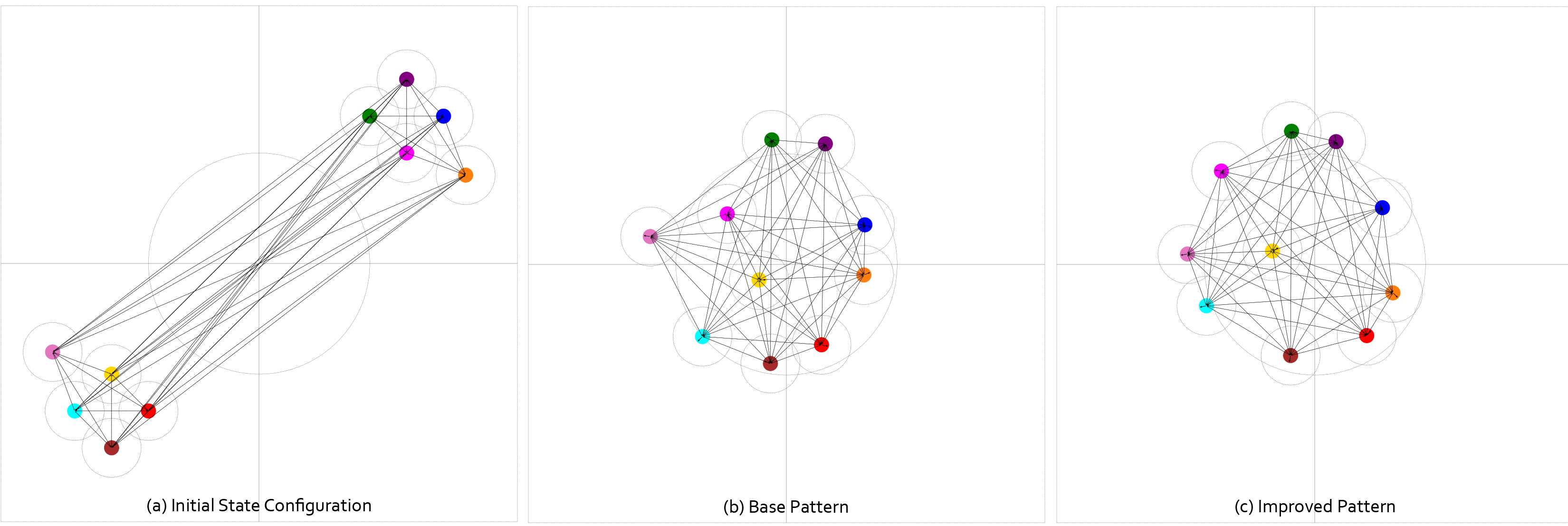}
		\caption{[Exp F] Pattern formation for swarm of size 10 in distributed initial state with undefined gathering point  }
		\label{fig:res10i}
	\end{figure}

\begin{figure}[!t]
	\vspace{-6mm}
		\centering
		\includegraphics[width=8cm]{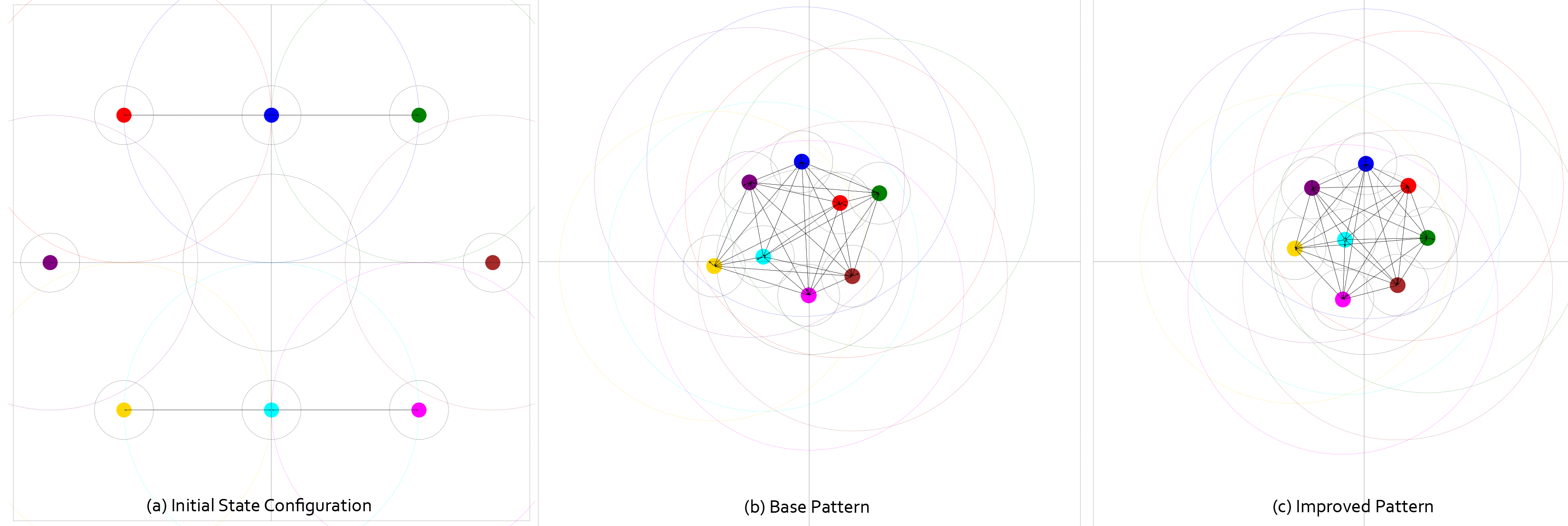}
		\caption{[Exp B] Pattern formation for swarm of size 8 in scattered initial state with pre-defined gathering point  }
		\label{fig:res8}
		\vspace{-6mm}
	\end{figure}

\begin{figure}[!t]
		\centering
		\includegraphics[width=8cm]{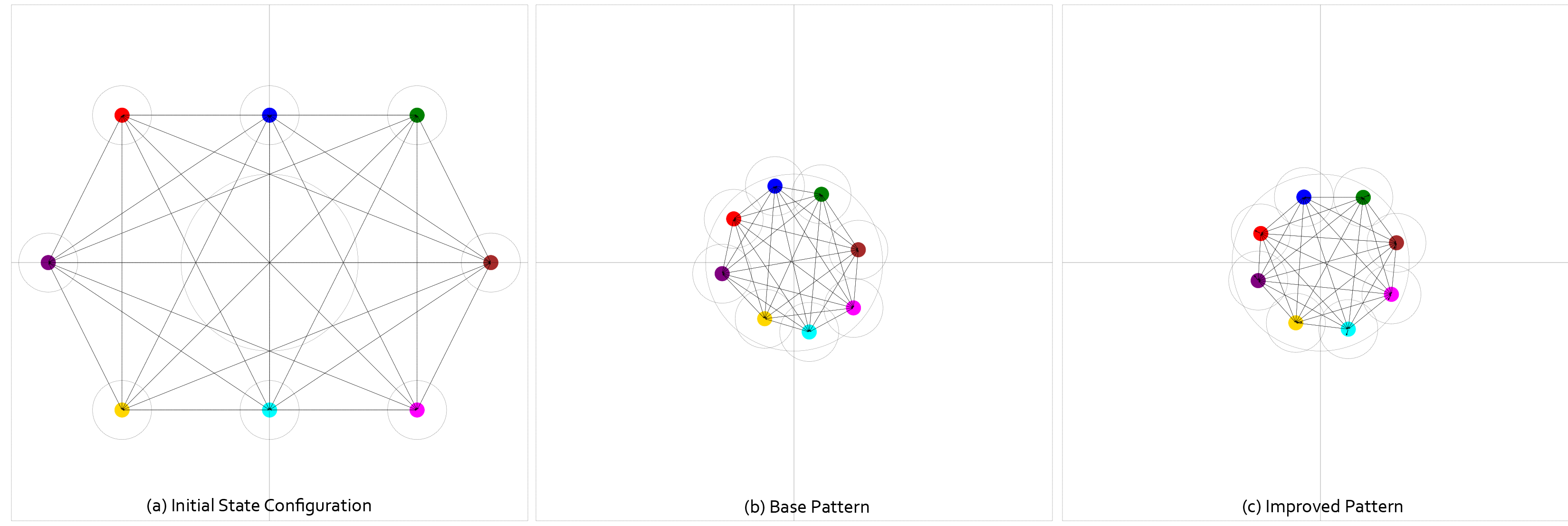}
		\caption{[Exp E] Pattern formation for swarm of size 8 in scattered initial state with undefined gathering point  }
		\label{fig:res8i}
	\end{figure}

	The patterns discovered by the trained policies are shown in Figs. \ref{fig:res10}--\ref{fig:res8i}, which show the (a) initial states, (b) the final pattern formed by base policy, and (c) the improved pattern formed by auxiliary policy.  It should be noted that the patterns discovered are quite different in both the cases and the proposed algorithm is able to gather robots even when gathering  without a predefined point. Similar results can be seen for swarm size of 8 (in experiments B and E) in Figs.~\ref{fig:res10i} and \ref{fig:res8i} which also show that the auxiliary policy is able to improve the base pattern even though the improved pattern is only slightly different in case of experiment E as seen in Fig.~ \ref{fig:res8i}.
 
	\begin{figure}[!t]
	\vspace{-2mm}
		\centering
		\includegraphics[width=8cm]{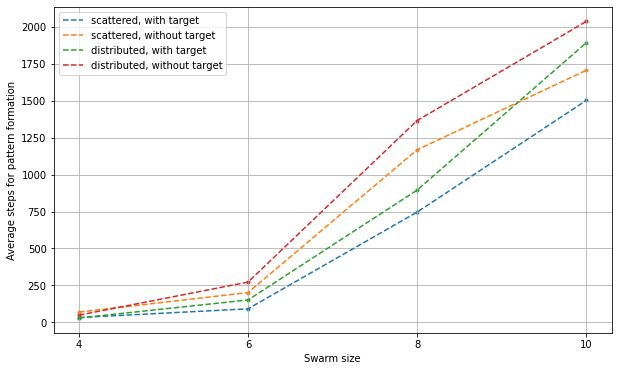}
		\caption{Swarm size vs average pattern formation steps for distributed and scattered initial state configurations }
		\label{fig:reln}
	\end{figure}

	\begin{table}[!t]
		\begin{center}
			\caption{Average Steps taken for pattern formation}
			\begin{tabular}{ |c|c|c|c|c| } 
				\hline
				Experiment & Swarm Size & Base Steps & Auxiliary Steps & Total Steps\\ 
				\hline
				A & 6 & 113.63 & 4.01 & 117.64 \\ 
				\hline
				B & 8 & 728.37 & 18.22 & 746.59 \\ 
				\hline
				C & 10 & 1730.20 & 162.38 & 1892.58 \\ 
				\hline
				
				D & 6 & 269.88 & 2.89 & 272.77 \\ 
				\hline
				E & 8 & 521.71 & 223.56 & 1168.83 \\ 
				\hline
				F & 10 & 1550.26 & 486.97 & 2037.23 \\ 
				\hline
			\end{tabular}
			\label{tab:avg_steps}
		\end{center}
	\end{table}
	
	Table \ref{tab:avg_steps} shows the average number of time-steps taken for robots to form a collision-less gathering pattern under different experimental settings and standard initial state configurations with uniform noise $\sim \mathcal{U}(-2R_{bot}, 2R_{bot})$. A total of $200$ random initial states were generated to test pattern formation steps under the corresponding initial state configuration. To study the relationship between swarm size and number of steps needed for pattern formation, we conducted similar experiments with robot swarms of sizes ranging from $4$ to $10$ under distributed and scattered initial state configurations; see Fig.~\ref{fig:reln}. Note that the number of steps varies almost linearly with the swarm size.

	\section{Conclusions and Future Work}
    In this paper, we proposed a DRL-based framework for automatic pattern discovery and gathering for swarms of oblivious fat opaque robots. Using state-of-the-art DRL techniques for policy optimization and proper reward system design, the agents trained using the proposed methods were shown to be able to discover collision-less trajectories for gathering in desirable patterns. With the use of multiple policies at different stages of gathering, it was possible for the robots to improve their patterns successively instead of being struck at local optima. 
    	
    In the future, we shall work upon improving the robustness and scale of our models by making them fully distributed and autonomous while also being able to handle environmental uncertainty like unseen obstacles, actuator failures, and recovery mechanisms. We look forward to explore trajectory planing and environmental mapping as well.

\pagebreak	
		
\end{document}